# SUB-IMAGE HISTOGRAM EQUALIZATION USING COOT OPTIMIZATION ALGORITHM FOR SEGMENTATION AND PARAMETER SELECTION

Emre Can Kuran[1], Umut Kuran[2] and Mehmet Bilal Er[2]

[1]Department of Software Engineering,
Bandırma Onyedi Eylül University, Balıkesir, Turkey
[2]Department of Computer Engineering, Harran University, Şanlıurfa, Turkey

## ABSTRACT

*Contrast enhancement is very important in terms of assessing images in an objective way. Contrast enhancement is also significant for various algorithms including supervised and unsupervised algorithms for accurate classification of samples. Some contrast enhancement algorithms solve this problem by addressing the low contrast issue. Mean and variance based sub-image histogram equalization (MVSIHE) algorithm is one of these contrast enhancements methods proposed in the literature. It has different parameters which need to be tuned in order to achieve optimum results. With this motivation, in this study, we employed one of the most recent optimization algorithms, namely, coot optimization algorithm (COA) for selecting appropriate parameters for the MVSIHE algorithm. Blind/referenceless image spatial quality evaluator (BRISQUE) and natural image quality evaluator (NIQE) metrics are used for evaluating fitness of the coot swarm population. The results show that the proposed method can be used in the field of biomedical image processing.*

## 1. INTRODUCTION

There are factors that affect the quality of the image such as contrast, noise and illumination. Contrast is the difference between darkest and brightest parts belong to an image. Hence, higher contrast makes the image regions more separable and it is important to enhance the distorted contrast [1]. Contrast enhancement is significant in the field of computer vision and used for several applications such as retinal image enhancement [2], underwater image enhancement [3] and chest x-ray enhancement [4]. Researchers proposed many techniques for enhancing the contrast of images. Histogram equalization (HE) is one of the most simplest methods proposed [5]. It simply increases the dynamic range of an image by redistributing pixel intensities. For this purpose, HE makes use of probability density functions (PDF) and cumulative distribution functions (CDF). It first computes the image histogram. After calculating the values for PDF and CDF functions according to the histogram of the image, it applies transformation on the output image using the appropriate transformation function. Different HE methods based on the classical HE is proposed in order to overcome limitations of the HE. Brightness preserving bi-histogram equalization (BBHE) separates image histogram into two parts according to the mean of the histogram and equalizes these parts independently [6]. Equal area dualistic sub-image histogram equalization (DSIHE) does not make use of mean but median using the same methods as BBHE [7]. Recursive sub-image histogram equalization (RSIHE) also uses median of the histogram, but it continues to separate image histogram until achieving a certain recursive level [8]. Exposure based sub-image histogram equalization (ESIHE) considers bright and dark regions separately for HE [9]. Mean and variance based sub-image histogram equalization (MVSIHE) [10] divides image histogram into four regions according to mean and variance difference, and employs a delta parameter to fuse input image and output image.

Algorithms like HE, BBHE, DSIHE, RSIHE and ESIHE suffers from different artifacts that occur in the output image but MVSIHE is one of the most outstanding methods among other HE techniques according to [11], [12]. Although it has a relatively good performance, its performance mainly depends on the delta parameter, which controls the rate of the image fusing [10]. Since contrast enhancement algorithms are widely employed in the field of medical imaging and biomedical image processing [13]–[15], we tried to provide optimal image quality for the knee x-ray images in this study. We have selected delta parameter and segmentation thresholds of MVSIHE algorithm using coot optimization algorithm (COA) [16]. Metaheuristic algorithms try to minimize/maximize value of the determined fitness function with respect to the problem type. For the fitness function, we have employed Blind/referenceless image spatial quality evaluator (BRISQUE) [17] and Natural Image Quality Evaluator (NIQE) [18] which are used for measuring image quality. These metrics are robust and insensitive to changes, since they are trained on wide variety of images. Besides that they are capable of assessing different kinds of distortions in the image. COA is also another novel optimization algorithm with a good performance as it is claimed in the original study. As it is pointed out in [11], MVSIHE preserves main brightness of the resultant image and also does not cause artifacts, however, we can't ensure that we found the optimum solution. Hence, we focused on improving its performance via parameter selection using COA. Rest of this paper is organized as follows. In Section 2, the used materials and methods are explained. In Section 3, experimental results are given and the proposed method is discussed with its advantages and disadvantages. In Section 3, a conclusion is made.

## 2. MATERIALS & METHODS

### 2.1. Coot Optimization Algorithm

The COA is a novel optimization algorithm proposed in [16], which is inspired from behavior of the coot birds. COA tries to simulate collective behaviors of the coots. The coots are directed by a few coots on the water surface. They have four distinct behaviors from observations: random movement, chain movement, position adjusting with respect to the group leaders and leading the group towards optimal area. We need a mathematical model to implement these behaviors.

First of all, a random population of coots is generated at the beginning. Assume that we have a multi-dimensional problem need to be solved for D dimensions, a population of N coots can be generated using Equation 1.

$$\text{PosCoot}(i) = \text{random}(1, D) \times (UB-LB) + LB, \quad i=1, 2, \ldots, N \qquad (1)$$

In Equation 1, the position of the coots in multi-dimensional space is generated randomly, with respect to the upper bounds UB and lower bounds LB that determined for each dimension.

Hence, the coots are prevented to overflow or underflow these limits. This initial random population is also evaluated according to a selected fitness function given in Equation 2.

$$F(i) = \text{Fitness}(\text{PosCoot}(i)), \quad i = 1, 2, \ldots, N \tag{2}$$

In order to model random movement of coots, first, a random position is produced according to the Equation 3. As the second step, the new position of the coot is computed according to the Equation 4.

$$R = \text{random}(1, D) \times (UB - LB) + LB \tag{3}$$

$$\text{PosCoot}(i) = \text{PosCoot}(i) + A \times RN2 \times (R - \text{PosCoot}(i)) \tag{4}$$

In Equation 4, RN2 is a random number in the range of [0, 1]. A and B are determined according to the Equation 5:

$$A = 1 - \left(T(i) \times \frac{1}{\text{IterMax}}\right), \; B = 2 - \left(T(i) \times \frac{1}{\text{IterMax}}\right) \quad i = 1, 2, \ldots, \text{IterMax} \tag{5}$$

In Equation 5, T(i) is the current iteration, IterMax is the maximum number of iterations. In order to move a coot towards another coot to implement chain movement, average position of the two coots is employed as given in Equation 6.

$$\text{PosCoot}(i) = 0.5 \times (\text{PosCoot}(i-1) + \text{PosCoot}(i)) \tag{6}$$

Coots also select a leader coot and follow them using Equation 7:

$$L_{ind} = 1 + (i \, \text{MOD} \, N_L) \tag{7}$$

In Equation 7, $L_{ind}$ is the index of the leader and $N_L$ is the number of leaders that determined as a parameter. A probability p is also defined. Finally, the rules given in Equation 8 is employed for determining leader positions.

$$\text{LeaderPos}(i) = \begin{cases} B \times R3 \times \cos(2R\pi) \times (\text{gBest} - \text{LeaderPos}(i)) + \text{gBest} & R4 < P \\ B \times R3 \times \cos(2R\pi) \times (\text{gBest} - \text{LeaderPos}(i)) + \text{gBest} & R4 \geq P \end{cases} \tag{8}$$

In Equation 8, R3 and R4 are random numbers in the range of [0, 1], gBest is the current global best, π is 3.14. Pseudocode of the COA is given in Figure 1.

```
1    Initialize the first population of coots randomly by Equation 1
2    Initialize the termination criteria, probability p, number of leaders and
     number of coots
3    Ncoot=Number of coots-Number of leaders
4    Random selection of leaders from the coots
5    Calculate the fitness of coots and leaders
6    Find the best coot or leader as the global optimum while the end criterion is
     not satisfied
7      Calculate A, B parameters by Equation 5
8      If rand< P
8         R, R1, and R3 are random vectors along the dimensions of the problem
9      Else
10        R, R1, and R3 are random number
11     End
12     For i=1 to the number of the coots
13        Calculate the parameter of K by Equation 7
14        If rand>0.5
15           Update the position of the coot by Equation 8
16        Else
17           If rand<0.5 i~=1
18              Update the position of the coot by Eq 6
19           Else
20              Update the position of the coot by Eq 4
21           End
22        End
23        Calculate the fitness of coot
24        If the fitness of coot < the fitness of leader(k)
25        Temp=leader(k); leader(k)=coot; coot=Temp;
26        end
27     End
28     For number of Leaders
29         Update the position of the leader using the rules given in Equation 8
30         If the fitness of leader < gBest
31         Temp= gBest; gBest =leader; leader=Temp; (update global optimum)
32         end
33     End
34     Iter=iter+1;
35   end
36   Postprocess results
```

Figure 1. Pseudocode of the COA.

## 2.2. Mean and Variance based Sub-Image Histogram Equalization Algorithm

MVSIHE algorithm can be divided into 5 stages in the following order: histogram segmentation, histogram bin modification, histogram equalization, normalization and image fusing.

### 2.2.1. Histogram Segmentation

Firstly, input image histogram is divided into two sub-histograms using a threshold k. The probability density function (PDF) of these two parts are computed. Then, for the first separation level k, two variables namely $\omega_0$ and $\omega_1$ can be given as in Equation 9.

$$\omega_0 = \sum_{i=0}^{k} PDF(i), \quad \omega_1 = \sum_{i=k+1}^{I_{max}} PDF(i) \qquad (9)$$

In Equation 8, i is the processed intensity level and $I_{max}$ is the maximum intensity level that is possible (256 for 8-bit). Mean of each part $\mu_0$ and $\mu_1$ can be given as in Equation 10.

$$\mu_0 = \sum_{i=0}^{k} PDF(i), \quad \mu_1 = \sum_{i=0}^{I_{max}} PDF(i) \tag{10}$$

In Equation 9, $I_{max}$ is the maximum intensity level and i is the intensity level. Whole image mean can be given as in Equation 11 and variance of the two parts can be defined by Equation 12, respectively.

$$\mu_1 = \mu_0 \omega_0 + \mu_1 \omega_1 \tag{11}$$

$$\sigma^2(k) = \omega_0 (\mu_0 - \mu)^2 + \omega_1 (\mu_1 - \mu)^2 \tag{12}$$

The MVSIHE algorithm finds the maximum value of the variance $\sigma^2$. After finding optimum threshold $k_{opt}$ (or $k_{H2}$), same procedure from Equation 9 to Equation 12 is repeated for two distinct histograms. Threshold of the lower sub-histogram, namely, $k_{H1}$ and threshold of the upper sub-histogram, namely, $k_{H3}$, are employed to determine other separation points. Thus, the segmented histogram is given in Equation 13 with its four sub-histograms.

$$H[I_{lowb}, I_{upb}] = \bigcup_{i=1}^{4} sub^{i,4}[I_{lowb}, I_{upb}] \tag{13}$$

In Equation 13, $I_{lowb}$ is the lower bound intensity level, $I_{upb}$ is the upper bound intensity level for the four sub-histograms.

### 2.2.2. Histogram Bin Modification

First of all, PDF for the sub-histogram i can be expressed as in Equation 14.

$$PDF_{sub^{i,4}} = \frac{sub^{i,4}}{n_{i,4}} \tag{14}$$

In Equation 14, $n_{i,4}$ is the number of pixels in the sub-histogram i. MVSIHE applies a histogram bin modification to overcome the domination of high frequency intensity levels and to balance high frequency and low frequency intensity levels [16]. Histogram bin modification is given by Equation 15.

$$MODIFIED\_PDF_{sub^{i,4}} = \left( \frac{e^{PDF_{sub^{i,4}}} - e^{-PDF_{sub^{i,4}}}}{e^{PDF_{sub^{i,4}}} + e^{-PDF_{sub^{i,4}}}} \right) \tag{15}$$

In Equation 15, e is the exponential function. Cumulative distribution function (CDF) of each sub-histogram is then calculated using Equation 16.

$$CDF_{sub^{i,4}}(x) = \sum_{j=I^{i,4}_{lowb}}^{I^{i,4}_{upb}} MODIFIED\_PDF(j), \quad \text{for } x = I^{i,4}_{lowb}, \ldots, I^{i,4}_{upb} \tag{16}$$

### 2.2.3. Histogram Equalization

HE is applied to each sub-histogram separately instead of global HE. A transformation function which considers the upper and lower boundaries of the sub-histogram is used in this case. Hence, a scaled HE is made and each of the intensity levels is equalized in its own range. Equation 17 is employed for this purpose.

$$f_{sub^{i,4}}(x)= I^{i,4}_{lowb}+\left(I^{i,4}_{upb}- I^{i,4}_{lowb}\right)\times CDF_{sub^{i,4}}(x), \quad \text{for } x=I^{i,4}_{lowb}, \ldots, I^{i,4}_{upb} \quad (17)$$

After each sub-histogram is equalized, they are merged to generate the final image.

### 2.2.4. Normalization

Since the transformation is applied distinctly for each sub-histogram, brightness saturation and artifacts might occur when the distinct histograms are merged, due to non-uniform illumination. Therefore, a normalization is applied on the image according to the Equation 18:

$$T(X)=\frac{X-X_{min}}{X_{max}-X_{min}}\times\left(X_{upb}-X_{lowb}\right)+X_{lowb} \quad (18)$$

In Equation 18, X is the input image which is actually the output of Equation 17, $X_{min}$ is the minimum intensity level in X, $X_{max}$ is the maximum intensity level in X, $X_{upb}$ is the upper bound (which is 255 for 256 levels) and $X_{lowb}$ is the lower bound (0).

### 2.2.5. Image Fusing

The original input image and output of Equation 18, are fused in this stage with the aim of preserving more information in the final image. The δ parameter is selected for determining the fusing rate, which is in the range of [0, 1], it determines the domination of the input image to the resultant image and vice versa. The fusing is done according to Equation 19.

$$O=\delta\times I_N+(1-\delta)\times I \quad (19)$$

In Equation 19, O is the final image, $I_N$ is the normalized image obtained using Equation 18, and I is the input image, respectively.

### 2.3. Proposed Method

The proposed method consists of three stages as follows: Determination of COA parameters and defining fitness function, selection of MVSIHE parameters via COA and enhancing the image using the selected parameters. The optimum thresholds for segmentation ($k_{H1}$, $k_{H2}$, $k_{H3}$) and delta (δ) parameter are selected by COA which employs the fitness function given in Equation 26. The illustration of the proposed method is given in Figure 2.

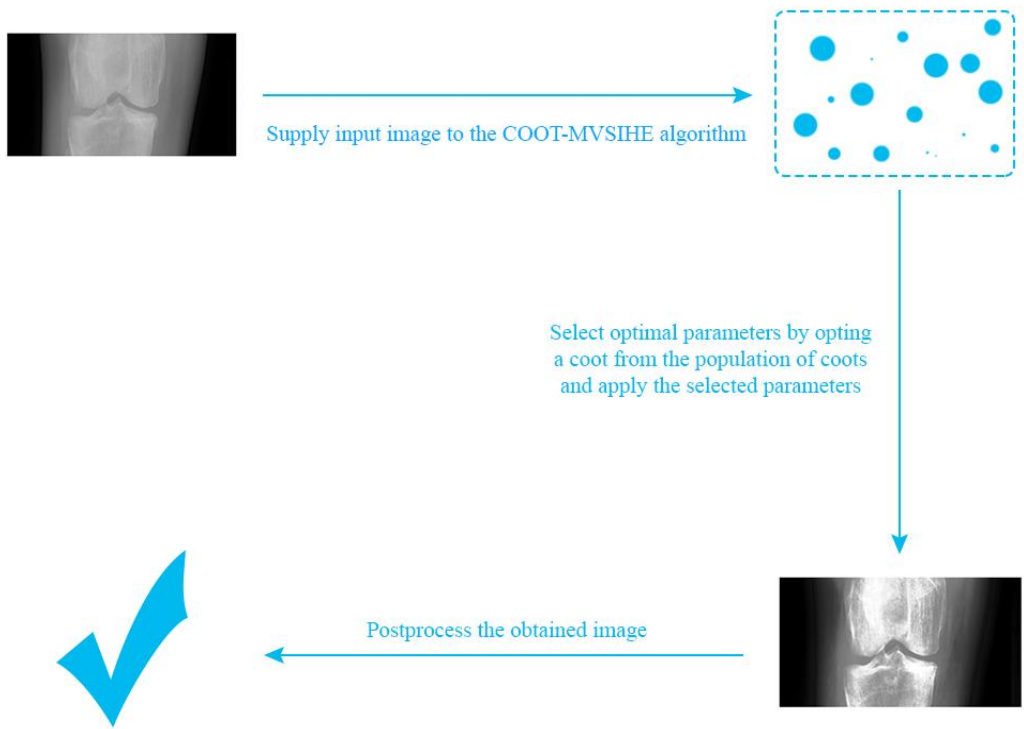

Figure 2. Illustration of the proposed method.

### 2.3.1. Determination of Coot Optimization Algorithm Parameters

Like in other metaheuristic optimization algorithms, parameters and fitness function of COA need to be determined. Parameters can be selected experimentally. In this study, we selected parameters to be reasonable as possible for an image enhancement task. We opted maximum number of iterations and population size as 10 to be relatively small as compared to some other optimization tasks in the literature since image enhancement is a long process that depends on the image size and bit depth. Upper bounds and lower bounds are specified in order to prevent COA to exceed limits. Number of dimensions is selected according to the problem type. Since we select delta parameter, and three histogram thresholds for MVSIHE, it is defined as four. The selected parameters are given in Table 1 for further information.

Table 1. Parameters of COA.

| Parameter (s) | Value (s) |
|---|---|
| Upper bounds | [0, 0, 51, 151] |
| Lower bounds | [1, 50, 150, 255] |
| Number of dimensions | 4 |
| Maximum number of iterations | 10 |
| Population size | 10 |
| Number of leaders | [Population Size * 0.1] |
| Probability | 0.5 |

**2.3.2. Fitness Function**

The aim of fitness function is to converge to the optimum solution. Each member of the population is evaluated using the determined fitness function with respect to the problem type. Because it is tried to enhance images using COA in this study, fitness function should be opted such that minimum distortion occurs in the resultant images. With this motivation, we selected BRISQUE and NIQE, which are developed to measure quality without needing any reference for various kind of distorted images.

BRISQUE [17] comprises three stages: extracting natural scene statistics (NSS), computing feature vectors and training support vector machines (SVM) [19] for predicting image quality scores. It is known that distribution of the distorted normalized images are different from the distribution of more natural normalized images. Distribution of the relatively less distorted images usually follow a bell curve therefore deviation from this curve can be perceived as a sign of distortion. Mean Subtracted Contrast Normalization (MSCN) is employed in BRISQUE in order to normalize an image. First, to calculate MSCN coefficients, the image is transformed to a luminance matrix as given in Equation 20:

$$\hat{I}(i,j) = \frac{I(i,j) - \mu(i,j)}{\sigma(i,j) + C}, \quad i=1, 2, \ldots, M, \ j=1, 2, \ldots, N \quad (20)$$

In Equation 20, I is the image, M and N are height and width of the image, i and j are spatial coordinates in x-axis and y axis, respectively. C is a small constant that is added for making sure the denominator is not equal to zero. The $\mu$ is the local mean field whereas $\sigma$ is the local variance field. Assume that GB is the Gaussian blur window (GBW) and I is the image, we can compute $\mu$ by using Equation 21 and $\sigma$ by employing Equation 22.

$$\mu = GB * I \quad (21)$$

$$\sigma = \sqrt{GB * (I - \mu)^2} \quad (22)$$

Since distortion also depends on the relationship of the pixels, pair-wise product of MSCN image with a shifted MSCN image is calculated. These pair-wise product images are horizontal, vertical, left-diagonal and right-diagonal. In the second step, the 5 images obtained in the first step (MSCN and pair-wise product images) are employed for feature extraction. MSCN image is fitted to a generalized Gaussian distribution (GGD) whereas pair-wise product images are fitted to an asymmetric generalized Gaussian distribution (AGGD). GGD has two parameters (shape and variance) and AGGD has four parameters (shape, mean, left variance and right variance). Hence, 2 features are obtained from MSCN image and 16 features (4 x 4) are obtained from pair-wise product images. The image is downsized by two (half of its original size), the same feature extraction technique is repeated and thus 36 features are gathered. In the last step of BRISQUE, feature vectors as inputs and their quality scores as outputs are fed to SVM and a model is trained. This model is used for predicting image quality score afterwards.

NIQE [18] consists of five phases: extracting NSS, patch selection, patch characterization, fitting patches to the multivariate Gaussian model (MGM) and applying NIQE index. For the first phase, NSS extraction, the process is similar to the NSS extraction in BRISQUE except that NIQE is only trained on the natural images but not distorted images. Hence, NIQE does not depend on any particular distortion type. In the second phase, the image is divided into $P \times P$ patches and patches exceeding a threshold T (which is defined as 0.75 in the original study) are selected according to the average local deviation field $\delta_L$ given in Equation 23.

$$\delta_L(t) = \sum\sum_{(i,j)\in patch} \sigma(i,j), \quad t=1, 2, ..., N_P \tag{23}$$

In Equation 23, i and j are the spatial coordinates belong to patch, t is the patch index, $N_P$ is the number of patches, and $\sigma(i, j)$ is the related variance value. In the third phase, similar to the BRISQUE, GGD and AGGD are employed for fitting and with a downsizing, a total number of 36 features are obtained. In the fourth phase, the NSS features obtained in the previous phases are fitted with an MGM model. MGM density can be given as in Equation 24.

$$f(x_1, ..., x_k) = \frac{1}{(2\pi)^{\frac{k}{2}}|\Sigma|^{\frac{1}{2}}} \exp\left(-\frac{1}{2}(x-v)^T\Sigma^{-1}(x-v)\right) \tag{24}$$

In Equation 24, $(x_1, ..., x_k)$ are the features, $\Sigma$ is the covariance matrix and v is the mean. In the last phase, the quality of the distorted images can be computed using Equation 25.

$$D(v_1,v_2,\Sigma_1,\Sigma_2) = \sqrt{\left((v_1-v_2)^T\left(\frac{\Sigma_1+\Sigma_2}{2}\right)^{-1}(v_1-v_2)\right)} \tag{25}$$

In Equation 25, $v_1$ and $v_2$ are the mean vectors whereas $\Sigma_1$ and $\Sigma_2$ are covariance matrices of natural MVG model and distorted image, respectively.

We have used BRISQUE and NIQE together by multiplying their outputs. The results of this multiplication is opted as the fitness function for a minimization problem. Because as the BRISQUE and NIQE scores increase, the quality of the evaluated images are decreased. We have used these metrics together to take advantage of the powerful sides of them. The preferred fitness function is defined in Equation 26.

$$F(I) = BRISQUE(I) \times NIQE(I) \tag{26}$$

In Equation 26, I is the input image given to the fitness function F.

## 2.4. Dataset and Preprocessing

Digital knee x-ray images [20] dataset is employed in the study for enhancing knee x-ray images. The images are obtained using PROTEC PRS 500E x-ray machine. This datasets includes images with labelled severity levels (using Kellgren and Lawrence grades) with the help of two distinct medical experts. In this study, we used the images contained in subfolder "MedicalExpert-I" for evaluating our method, since the images are same and only labels differ, the other subfolder is ignored. The images are first converted to gray before evaluating the proposed method. Further details about the dataset can be found in Table 2.

Table 2. Details of the dataset.

| Expert ID | Number of images for severity level | | | | | Total number of images | Bit depth | Types of Images |
|---|---|---|---|---|---|---|---|---|
| | Normal | Doubtful | Mild | Moderate | Severe | | | |
| I | 514 | 477 | 232 | 221 | 206 | 1650 | 8-bit | PNG |
| II | 503 | 488 | 232 | 221 | 206 | 1650 | 8-bit | PNG |

## 2.5. Evaluation Metrics

### 2.5.1. Absolute Mean Brightness Error

Absolute mean brightness error (AMBE) calculates the mean brightness error, hence, a lower AMBE indicates a better brightness preservation. AMBE is defined as in Equation 27.

$$AMBE = |M_I - M_O| \qquad (27)$$

In Equation 27, $M_I$ and $M_O$ are the mean values of the input image and the output image respectively. AMBE is defined in the range of [0,255].

### 2.5.2. Peak Signal-to-Noise Ratio

Peak signal-to-noise ratio (PSNR) is employed for measuring the error between input and output image after an operation that might present distortion. Hence, a higher value of PSNR indicates better image quality and less distortion. In order to compute PSNR, first of all, mean squared error (MSE) is computed using Equation 28. Then, PSNR is calculated using Equation 29.

$$MSE = \frac{1}{M \times N} \sum_{i=1}^{M} \sum_{j=1}^{N} (I(i,j) - O(i,j))^2 \qquad (28)$$

$$PSNR = 10 \log_{10} \frac{(2^n - 1)^2}{\sqrt{MSE}} \qquad (29)$$

In Equation 28, I and O are input image and output image, respectively, and i and j are the spatial coordinates of x-axis and y-axis, respectively. In Equation 29, n is determined according to image type, that is, for 256 gray levels, it is equal to 8 since $2^8$ is equal to 256. Higher PSNR value indicates better quality, which is infinite when MSE is equal to 0.

### 2.5.3. Structured Similarity Index

Structured similarity index (SSI) is a measures the similarity index by considering the input and output image. Higher SSI value indicates more similarity. SSI value is in the range of [-1, 1]. SSI could be calculated by employing Equation 30.

$$SSI = \frac{(2M_I M_O + C_1)(2\sigma_{IO} + C_2)}{(M_I^2 + M_O^2 + C_1)(\sigma_I^2 + \sigma_O^2 + C_2)} \qquad (30)$$

In Equation 30, $M_I$, $M_O$, $\sigma_I$, $\sigma_O$, $\sigma_{IO}$, $\sigma_{IO}$, $C_1$ and $C_2$ are mean value of the input image, mean value of the output image, standard deviation of the input image, standard deviation of the output image, first constant and second constant, respectively. The constants are usually selected as small values such that they are close to zero which ensures that numerator and denominator to be greater than zero.

## 3. RESULTS & DISCUSSION

In this section, visual results, performance evaluation results, stability results and convergence curves are provided. Visual results are important since performance metrics are not enough to compare image enhancement algorithms. Performance evaluation results are also given for an objective assessment. Stability results are provided because the results (outputs) of the

Computer Science & Information Technology (CS & IT) 43

metaheuristic algorithms may change slightly at each run and convergence curves are given to indicate performance of the COA algorithm on the dataset.

## 3.1. Performance Evaluation Results

The performance evaluation results are given in Table 3 in terms of AMBE, PSNR and SSI. First best results are in bold whereas second best results are underlined. The evaluation results show that our method is capable of enhancing different classes of images in the dataset. Proposed method outperforms most of the other methods compared in this study and exhibits competitive performance as compared to some of them. Although, visual results in Section 3.2 indicate that our method is better at preserving details in the image.

Table 3. Average performance evaluation results for AMBE, PSNR and SSI on 1650 knee x-ray images.

| Method | AMBE | PSNR | SSI |
|---|---|---|---|
| HE | 22.5258 | 16.6345 | 0.6898 |
| BBHE | 21.4886 | 17.5533 | 0.7266 |
| DSIHE | 18.4800 | 17.3079 | 0.7021 |
| ESIHE | 9.8640 | 21.1487 | 0.8525 |
| MVSIHE(0.6) | **5.3629** | 24.6394 | **0.9185** |
| COA-MVSIHE | 7.9919 | **26.2748** | 0.9139 |

## 3.2. Visual Results

Visual results for the images belong to distinct classes (severity levels) are given in Figure 3 for compared methods. The proposed method provides a balanced enhancement and prevents most of the detail loss as it can be observed from the figure. For example, the enhanced output of the sample output image that belong to the mild class provides more details and higher contrast as compared to the outputs of other methods. Other outputs are also balanced and not too sharp or washed-out (saturated).

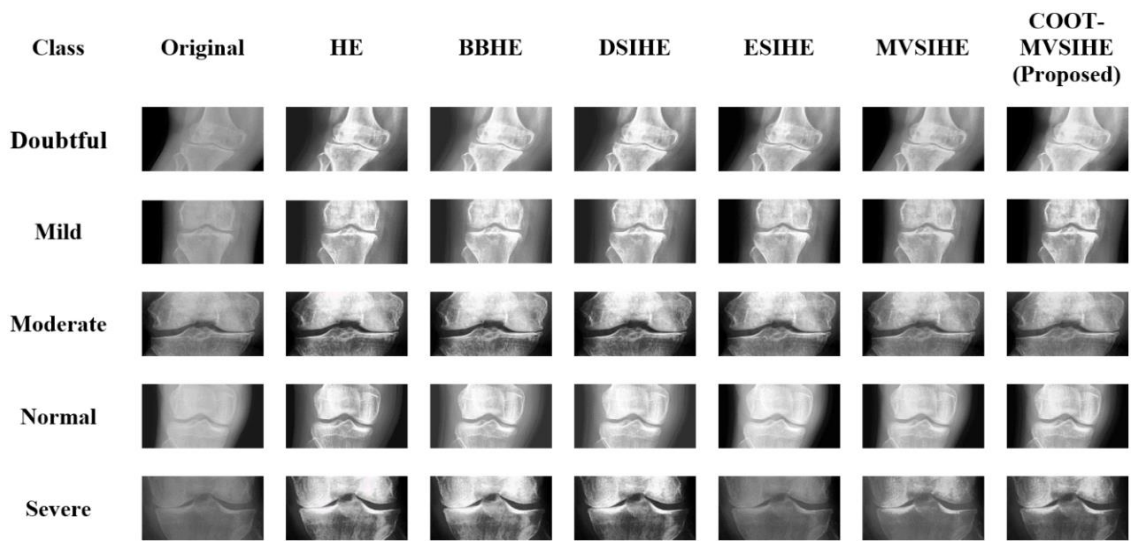

Figure 3. Visual results for images belong to different severity levels (classes) and algorithms.

## 3.3. Convergence Curves and Stability Results

Convergence curves of the COA algorithm employed for the image enhancement task are given in Figure 4. As it can be seen from the convergence curves that as the number of iterations increases, the COA converges to the minimum (optimum value) better. We can conclude that increasing the maximum number of iterations might increase the chance of finding global optimum.

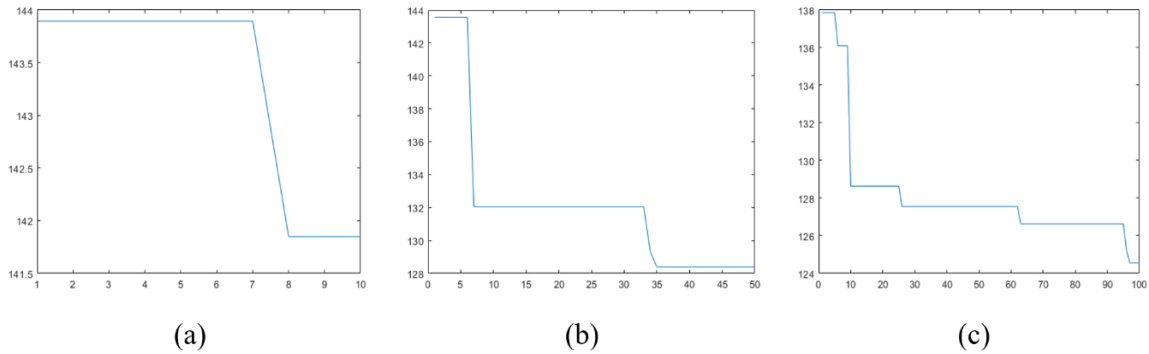

(a)　　　　　　　　　　　　(b)　　　　　　　　　　　　(c)

Figure 4. Convergence curves for different number of iterations with the same population size of 10: (a) 10 iterations. (b) 50 iterations. (c) 100 iterations.

Since COA is a metaheuristic optimization algorithm, different results might be obtained at different runs. In this study, the stability of the proposed method is tested using 10 sample images by performing 10 runs. The parameters given in Table 1 is used for the test. The stability results are given in Table 4. We used variance, standard deviation and range (difference between maximum and minimum value) for such aim. The standard deviation and variance values is lower than one and close to the zero for AMBE and PSNR. SSI varies relatively higher as compared to the other metrics. Since these metrics show variation of the data, these given results should be considered for the use of proposed method in further studies.

Table 4. Stability results of the proposed method for 10 samples and 10 runs.

| Measurement | AMBE | PSNR | SSI |
| --- | --- | --- | --- |
| Variance | 0,793405 | 0,000125 | 2,836652 |
| Standard Deviation | 0,845023 | 0,010618 | 1,597807 |
| Range | 2,56 | 0,0357 | 4,8175 |

## 3.4. Discussion

The proposed method is adaptive, that is, it can find the optimal parameters for the enhancement of wide variety of images. It does not need any pretrained structure. Its parameters could be changed according to device type for gaining more performance (such as decreasing number of iterations etc.). However, its stability should be considered when a more stable enhancement technique is significant. Also, changing parameters may increase or decrease the success of the enhancement task with respect to the data type.

## 4. CONCLUSIONS

In this study, a method combining COA with MVSIHE is proposed for enhancement of the knee x-ray images. The experimental results show that our method outperforms most of the other cutting-edge methods or at least shows a competitive performance. Visual results also indicate that our method provides a balanced enhancement for most of the images in the used dataset. Hence, the proposed method can support various supervised models by performing preprocessing task, which might be employed to increase the classification success in further studies.

46          Computer Science & Information Technology (CS & IT)

## AUTHORS

**Emre Can Kuran** was born in İzmir, Turkey, in 1997. He received the B.S. degree in computer engineering from Harran University, Turkey, in 2020, and currently is a M.Sc. student in computer engineering at Harran University, Turkey. He is currently working as a research assistant in Software Engineering, Bandırma Onyedi Eylül University, Turkey. His research interests include artificial intelligence, image processing, computer vision, data science, machine learning and software development.

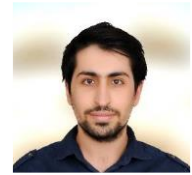

**Umut Kuran** was born in Şanlıurfa, Turkey, in 1982. He received the B.S. degree in mathematics from Harran University, Turkey, in 2005, the M.Sc. degree in mathematics from Celal Bayar University, Turkey, in 2007, and in computer science from Rutgers, The State University of New Jersey, USA, in 2013. He is currently a Ph.D. student in mathematics, at Harran University. He is currently working as a full time lecturer in Computer Engineering, Harran University, Turkey. His research interests include artificial intelligence, image processing, neural networks, fuzzy logic, numerical computation and cryptology.

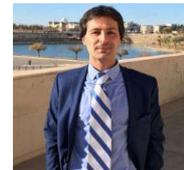

**Mehmet Bilal Er** was born in Şanlıurfa, Turkey, in 1988. He received the B.S. degree in computer engineering from Eastern Mediterranean University, Cyprus, in 2010, the M.Sc. degree in computer engineering from Cankaya University, Turkey, in 2013, and the Ph.D. degree in computer engineering from Maltepe University, Turkey, in 2019. He is currently an Assistant Professor with the School of Computer Engineering, Harran University, Turkey. His research interests include sound processing, pattern recognition, machine learning and deep learning.

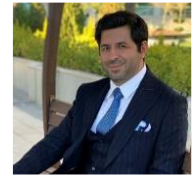